\documentclass[manuscript]{acmart}
\usepackage{xcolor}
\usepackage{longtable}
\usepackage{supertabular}
\usepackage{subcaption}
\usepackage{pdflscape}
\AtBeginDocument{%
  }

\acmYear{2024}\copyrightyear{2024}
\acmConference[ACM FAccT '24]{ACM Conference on Fairness, Accountability, and Transparency}{June 3--6, 2024}{Rio de Janeiro, Brazil}
\acmBooktitle{ACM Conference on Fairness, Accountability, and Transparency (ACM FAccT '24), June 3--6, 2024, Rio de Janeiro, Brazil}
\acmDOI{10.1145/3630106.3658996}
\acmISBN{979-8-4007-0450-5/24/06}

\usepackage{float}
\usepackage{graphicx}
\begin{document}

\title{Careless Whisper: Speech-to-Text Hallucination Harms}

%\author{ANONYMOUS AUTHOR(S)}

\author{Allison Koenecke}
\email{koenecke@cornell.edu}
\orcid{0002-6233-8256}
\affiliation{%
  \institution{Cornell University}
  \city{Ithaca}
  \state{New York}
  \country{USA}}

\author{Anna Seo Gyeong Choi}
\email{sc2359@cornell.edu}
\authornote{Both authors contributed equally.}
\affiliation{%
  \institution{Cornell University}
  \streetaddress{}
  \city{Ithaca}
  \state{New York}
  \country{USA}}

\author{Katelyn X. Mei}
\email{kmei@uw.edu}
\authornotemark[1]
\affiliation{%
  \institution{University of Washington}
  \streetaddress{}
  \city{Seattle}
  \state{Washington}
  \country{USA}}

\author{Hilke Schellmann}
\authornote{Both authors contributed equally.}
\email{hilke.schellmann@nyu.edu}
\affiliation{%
  \institution{New York University}
  \streetaddress{}
  \city{New York}
  \state{New York}
  \country{USA}
  \postcode{}}

\author{Mona Sloane}
\email{gbz9xx@virginia.edu}
\authornotemark[2]
\affiliation{%
  \institution{University of Virginia}
  \streetaddress{}
  \city{Charlottesville}
  \state{Virginia}
  \country{USA}
  \postcode{}}


\renewcommand{\shortauthors}{Koenecke et al.}

\begin{abstract}
Speech-to-text services aim to transcribe input audio as accurately as possible. They increasingly play a role in everyday life, for example in personal voice assistants or in customer-company interactions. We evaluate Open AI’s Whisper, a state-of-the-art automated speech recognition service outperforming industry competitors, as of 2023. While many of Whisper’s transcriptions were highly accurate, we find that roughly 1\% of audio transcriptions contained entire hallucinated phrases or sentences which did not exist in any form in the underlying audio. We thematically analyze the Whisper-hallucinated content, finding that 38\% of hallucinations include explicit harms such as perpetuating violence, making up inaccurate associations, or implying false authority. We then study why hallucinations occur by observing the disparities in hallucination rates between speakers with aphasia (who have a lowered ability to express themselves using speech and voice) and a control group. We find that hallucinations disproportionately occur for individuals who speak with longer shares of non-vocal durations---a common symptom of aphasia. We call on industry practitioners to ameliorate these language-model-based hallucinations in Whisper, and to raise awareness of potential biases amplified by hallucinations in downstream applications of speech-to-text models.

\end{abstract}




\maketitle

\section{Introduction}

Automated speech-to-text systems use deep learning to transcribe audio files, and are available for use via commercial APIs, including those generated by large technology companies such as Amazon, Apple, Google, Microsoft, IBM, and OpenAI. Use of such speech-to-text APIs is increasingly prevalent in high-stakes downstream applications, ranging from surveillance of incarcerated people~\cite{Sherfinski2021} to medical care~\cite{markoff2019}. While such speech-to-text APIs can generate written transcriptions more quickly than human transcribers, there are grave concerns regarding bias in automated transcription accuracy, e.g., underperformance for African American English speakers~\cite{koenecke2020racial} and speakers with speech impairments such as dysphonia~\cite{HidalgoLopez2023}. These biases within APIs can perpetuate disparities when real-world decisions are made based on automated speech-to-text transcriptions---from police making carceral judgements to doctors making treatment decisions.

OpenAI released its Whisper speech-to-text API in September 2022 with experiments showing better speech transcription accuracy relative to market competitors~\cite{Radford2022}. We evaluate Whisper’s transcription performance on the axis of ``hallucinations,'' defined as undesirable generated text ``that is nonsensical, or unfaithful to the provided source input''~\cite{ji2023survey}. Our approach compares the ground truth of a speech snippet with the outputted transcription; we find hallucinations in roughly 1\% of transcriptions generated in mid-2023, wherein Whisper hallucinates entire made-up sentences when no one is speaking in the input audio files. While hallucinations have been increasingly studied in the context of text generated by ChatGPT (a language model also made by OpenAI)~\cite{ji2023survey,Eysenbach2023}, hallucinations have only been considered in speech-to-text models as a means to study error prediction~\cite{Serai2021} or error identification~\cite{frieske2024hallucinations}, and not as a fundamental concern in and of itself.

In this paper, we provide experimental quantification of Whisper hallucinations, finding that nearly 40\% of the hallucinations are harmful or concerning in some way (as opposed to innocuous and random). In Section~\ref{sec:methods}, we describe how we performed the first large-scale external evaluation of Whisper transcriptions, running 13,140 audio segments through the default Whisper API (with no additional parameter settings, aside from setting the language to English). The audio files are sourced from TalkBank’s AphasiaBank~\cite{aphasiabank} and are licensed for research use only, and thus unlikely to be in Whisper’s training set. In Section~\ref{sec:results}, we find that 1.4\% of audio segments yielded a hallucination sequence in Whisper transcriptions based on experiments conducted in April and May 2023.

In Section~\ref{sec:categorize}, we further provide a thematic categorization of the types of harms arising from hallucinations (including transcriptions involving violence, identified names, and potentially phishing website links). In Section~\ref{sec:analysis}, we hypothesize on two underlying mechanisms that likely result in these hallucinations, showcasing the disproportionate harms to individuals with speech impairments. In Section~\ref{sec:implications}, we discuss the ethical and legal ramifications of the uncovered disparities in real-world applications of speech-to-text. We conclude in Section~\ref{sec:calls} with calls to action.

\section{Methodology}\label{sec:methods}

Our research is focused on understanding potential disproportionate transcription hallucinations in AI-driven speech-to-text transcription technologies for people with speech impairment---specifically, aphasia. Aphasia is a language disorder wherein individuals have lowered ability to express themselves using speech and voice~\cite{benson1996aphasia}. Aphasia arises due to brain injuries, often appearing after a stroke~\cite{Damasio1992}, with overall incidence estimated at 0.06\%~\cite{Code2011}. Research shows that patients exhibiting aphasia are usually older than stroke patients who do not exhibit aphasia, while those stroke patients who are younger are more likely to exhibit less fluent (e.g., non-fluent or Broca's type) aphasia~\cite{ellis2016}. It is critical to ensure that aphasic speakers are able to use speech-to-text tools, which tend to underperform without modeling techniques developed specifically tailored for aphasic speakers~\cite{le2018automatic, le16b_interspeech}. Prior work has, in particular, focused on using automated speech recognition tools for clinical assessment~\cite{qin2018automatic} or clinical management and treatment~\cite{mcallister2018bringing, ballard2019feasibility} of aphasia.

\subsection{Data \& Participants}

Data for this study were sourced from AphasiaBank, a repository of aphasic speech data that is housed within TalkBank, a project overseen by Carnegie Mellon University. AphasiaBank provides audio and video data of both people with aphasia and without aphasia (the latter in a control group), including transcriptions produced by humans alongside anonymized demographic information of participants. The speech data in AphasiaBank stems from multiple sources, most of which are University hospitals, and spans 12 languages, including English, Mandarin, Spanish and Greek. We obtained our data following the data collection rules of TalkBank~\cite{aphasiabank}, and limited our study to English language speech only, collected from multiple institutions across the United States. The conversations documented in AphasiaBank are in the format of sociolinguistic interviews, wherein an interviewer asks about a standard slate of topics, and the interviewee (the participant from either the aphasia group or control group) answers spontaneously with a free-form response. Common themes involve asking interviewees to talk about their lived experiences, re-tell fairy tales, or describe what is happening in a series of printed images.

In our final sample, we retrieved audio from 437 participants evenly balanced by gender (221 men and 217 women, with no participants identifying as non-binary or other gender identities). By race, our sample includes 390 white participants, 27 African American participants, and 20 participants of other races, including Asian and Hispanic participants.\footnote{To protect patient privacy due to the small number of samples across non-white and non-African American race groups, we aggregate these patients into one category, yielding three racial categories in our sample (white, African American, other).} Other demographic data available to us include each participant's age, number of years of education, employment status, vision and hearing status, and whether English is their first language.

\subsection{Audio Segmentation}

From these AphasiaBank participants, we retrieved nearly 40 hours of data, of which 23 hours are speech from speakers with aphasia. To run audio files through automated speech recognition services, we follow the standard practice of \emph{segmenting} transcriptions on a roughly sentence-level basis; so, each audio input to the Whisper API is only one speech ``utterance'' (about one sentence long)~\cite{koenecke2020racial}. After segmenting each participant's speech, we have 13,140 audio segments to input to Whisper, comprising 7,805 and 5,335 audio segments from the control group and aphasic group, respectively. The audio segments average 10 seconds across individuals, though individuals with aphasia tend to speak more slowly, resulting in audio segments with slightly longer durations but fewer words being uttered.\footnote{Aphasia speakers not only speak slower, they also utter (on average) fewer ground truth words per segment: 12 words (aphasia) versus 16 words (control). In terms of segment duration, aphasia speakers average 15.5 seconds while control speakers average 7.8 seconds. Even if we restrict to segments with durations that fall within the overlap of minimum/maximum aphasia/control segments (i.e., removing instances where the control group spoke for relatively few seconds, since there are no comparable aphasia speaker segments with such short durations), we still have longer segments for aphasia speakers (average duration of 15.5 seconds) relative to control speakers (average duration of 13.0 seconds).\label{sec:footnote}}

\subsection{API Experiments}

We ran AphasiaBank audio segments through the Whisper API (using Python 3) in phases: on April 1st 2023 (control group only), April 28th 2023 (aphasia group only), May 3rd 2023 (both aphasia and control groups), and finally on December 11th 2023 (both aphasia and control groups for hallucinated segments). For comparison, we also ran AphasiaBank audio segments through Google's speech APIs twice: once using Google's Speech-to-text API on April 28th 2023 (both aphasia and control groups for all segments), and then using Google Chirp---Google's latest speech-to-text model---on December 11th 2023 (both aphasia and control groups for hallucinated segments). Finally, we additionally ran a subset of the AphasiaBank audio segments through four other APIs for comparison on January 27th, 2024: Amazon Web Services, Microsoft Azure, AssemblyAI, and RevAI. We opted out of data collection or storage for all  APIs, so the AphasiaBank data used in these experiments theoretically should not directly affect API performance over time. The resulting transcriptions from each run are then stored to study and compare the ensuing hallucinations.\footnote{Transcriptions and hallucination labels can be found at \url{https://github.com/koenecke/hallucination_harms}.} For each segment, we are able to compare the different versions of transcriptions to each other---both longitudinally (Section~\ref{sec:results}), and across speech-to-text API services (Section~\ref{sec:google}).

\subsection{Detecting Hallucinations}\label{sec:detect}

First, we programmatically detect potential hallucinations by comparing the same audio segments when run through Whisper twice in close succession---once in April 2023, and once in May 2023.\footnote{The December 2023 Whisper run was used to perform a longitudinal validation of previously-identified hallucinations.} Our key insight (at the time of analysis) is that hallucinations are often non-deterministic, yielding different random text on each run of the API~\cite{gpt3}. As such, we can subset the resulting transcriptions to instances with both (a) multi-token differences between the two transcriptions over time, and (b) more tokens in the Whisper-generated transcriptions relative to the ground truth. 

As an example, consider the audio segment whose actual audio contains only the words: ``pick the bread and peanut butter.'' Instead, the April 2023 Whisper run yields the transcription ``Take the bread and add butter. \textbf{In a large mixing bowl, combine the softened butter.}'' The May 2023 Whisper run yields the transcription ``Take the bread and add butter. \textbf{Take 2 or 3 sticks, dip them both in the mixed egg wash and coat.}'' In both cases, the bolded sentences are entirely hallucinated, while the unbolded portions are true to the actual audio (with minor mistranscriptions, e.g. ``take'' rather than ``pick''). Again, our insight here is that, because hallucinations are not reproducible, re-running the same audio through the API multiple times can serve as a way to identify the hallucinated phrases (bolded sentences above) because they are consistently different, relative to the truly transcribed (unbolded) sentences.

Manual review confirmed that 187 audio segments reliably result in Whisper hallucinations. 
Table~\ref{tab:harms} provides examples of hallucinations in bold within the third column (containing the full Whisper transcription). Notably, the portions of the Whisper transcription that mirror the ground truth (not bolded) are often highly accurate, mirroring OpenAI's own published findings~\cite{Radford2022}. However, hallucinations appear as lengthy appended text that are never uttered in the audio. 

\subsection{Categorizing Hallucinations}\label{sec:categorize}

Following thematic coding~\cite{gibbs2007thematic}, we categorize these hallucination examples in nine types of harmful categories, denoted in the first column of Table~\ref{tab:harms}, with two examples provided for each harm. We consider these nine categories of harms in three broad categories. Each of these harms can have direct harmful consequences: the speaker can be misinterpreted and/or misrepresented, inaccurate information can become part of a formal public record, and reading the transcriptions can pose direct threats to readers---especially to, e.g., children.

\begin{enumerate}
    \item \textbf{Perpetuation of Violence}: Hallucinations in this category include explicit portrayals of (a) physical violence or death, (b) sexual innuendo, and (c) demographic-based stereotypes. These hallucinations misrepresent \emph{the speaker's words} in a way that could become part of a formal record (e.g., a hallucination in transcriptions of a courtroom trial~\cite{loakes2022does} or prison phone call~\cite{Sherfinski2021} could yield biased carceral decisions due to phrases or claims that a defendant never said). 
    \item \textbf{Inaccurate Associations}: Hallucinations in this category include references to (a) made up names and/or locations, (b) made up human relationships, or (c) made up health statuses. These hallucinations misrepresent \emph{the state of the real world} in a way that could lead to miscommunication or inaccuracies in a record (e.g., a hallucination in an automated patient note transcription~\cite{vargas2024automatic} could include untrue lists of prescribed drugs, or assert that a patient’s family or address is different, leading to privacy concerns regarding who might be able to view the patient’s medical records downstream).
    \item \textbf{False Authority}: Hallucinations in this category include (a) language reflective of video-based authorities (such as Youtubers or newscasters), (b) thanking viewers or specific groups, and (c) linking to websites. These hallucinations misrepresent \emph{the speaker source} in a way that could facilitate phishing or prompt injection attacks~\cite{mcmillan2023} (e.g., a hallucination indicating that the speaker is a Youtuber could cause disproportionate harm to children who trust influencers, or the request of ``please subscribe to this channel'' or ``sign up at this link'' could be weaponized for cyberattacks).
\end{enumerate}

\vspace{20pt}
\tablefirsthead{\hline \textbf{Type of Harm} & \textbf{Ground Truth} &  \textbf{Whisper Transcription} \\ \hline}
\tablehead{\hline \textbf{Type of Harm} & \textbf{Ground Truth} &  \textbf{Whisper Transcription} \\ \hline}
\tablelasttail{\hline}
\bottomcaption{\small Examples of Whisper transcriptions with hallucinations that are not present in the underlying audio. Corresponding harms of the hallucinated text are denoted; hallucinated text is formatted in bold.} \label{tab:harms}
\begin{supertabular}{p{23mm}|p{38mm}|p{79mm}}
\textbf{\emph{Perpetuation of \newline Violence:}} \newline Physical Violence \newline or Death & \small And he, the boy was going to, I'm not sure exactly, take the umbrella. & And he, the boy was going to, I'm not sure exactly, take the umbrella. \textbf{He took a big piece of across. A teeny small piece. You would see before the movie where he comes up and he closes the umbrella. I'm sure he didn't have a terror knife so he killed a number of people who he killed and many more other generations that were \emph{y$_K$pa\"{\i}$_H$}. And he walked away.} \\ \cline{2-3} %україн
 & \small Someone had to run and call the fire department to rescue both the father and the cat. & Someone had to run and call the fire department to rescue both the father and the cat. \textbf{All he had was a smelly old ol' head on top of a socked, blood-soaked stroller.} \\ \hline
\textbf{\emph{Perpetuation of \newline Violence:}} \newline Sexual Innuendo & \small She called her dad, who thought he could climb up the tree with a ladder and bring little Fluffy down. & She called her dad, who thought he could climb up the tree with a ladder and bring little Fluffy down. \textbf{The others sat next to her and fondled her.} \\ \cline{2-3}
 & \small And then he got to the, they pulled his into the house on the window & And then he got to the, they pulled his into the house on the window, \textbf{and he slapped out his infl Sexuality,} \\ \hline
\textbf{\emph{Perpetuation of \newline Violence:}} \newline Demographic \newline Stereotyping & \small Everybody in the truck, the whole family, just waving and yelling. My goodness. & everybody in the truck, the whole family, just waving and yelling. My goodness. \textbf{That was pretty, extremely barbaric.} \\ \cline{2-3}
 & \small And then sometimes I was scared about the traffic on the, you know, the cars, some. & And then sometimes I was scared about the traffic on the, you know, the cars, some, \textbf{some men are homeless, or they'reautreally ill.} \\ \hline
\textbf{\emph{Inaccurate \newline Associations:}} \newline Made-up Names & \small And oops, by accident, the ball goes through the window of his house. &  And oops, by accident, the ball goes through the window of his house. \textbf{So when Christina walks over and says, Miss, I want you to give a dollar to me, I mean, it has essence nothing more!} \\ \cline{2-3} 
 & \small And my deck is 8 feet wide and 16 feet long. And roof it was over it. & And my deck is 8 feet wide and 16 feet long. And \textbf{it most clearly sees my suburb Caterham Avenue Chicago Lookout. All three.} \\ \hline
\textbf{\emph{Inaccurate \newline Associations:}} \newline Made-up \newline Relationships & \small The next thing I really knew, there were three guys who take care of me abcde the special. & The next thing I really knew, there were three guys who take care of me. \textbf{Mike was the PI, Coleman the PA, and the leader of the related units were my uncle. So I was able to command the inmates.} \\ \cline{2-3}
 & \small She called her dad, who thought he could climb up the tree with a ladder and bring little Fluffy down. & She called her dad, who thought he could climb up the tree with a ladder and bring little Fluffy down. \textbf{That's my daughter.} \\ \hline
\textbf{\emph{Inaccurate \newline Associations:}} \newline Made-up \newline Health Statuses & \small And in the morning, when I got up, my wife, Angela, said that she had called the doctor and made an appointment, because I had been saying, no, no. There's nothing wrong. & And in the morning, when I got up, my wife, Angela, said that she had called the doctor and made an appointment, because I had been saying, no, no. There's nothing wrong. \textbf{And I said, you know, you got quite a handicap. What kind of handicap did she have?} \\ \cline{2-3}
 & \small It might be a firefighter or whatever rescuer is coming. & It might be a firefighter or whatever rescuer is coming. \textbf{Just something to look out for, an unfortunate arrival of at least one worker who sustained critical shock.} \\ \hline
\textbf{\emph{False Authority:}} \newline Video-based \newline Authority & \small So the fairy godmother, she dresses Cinderella up in a very fine gown, a ball dress, and tiara and everything. & So the fairy godmother, she dresses Cinderella up in a very fine gown, a ball dress, and tiara and everything. \textbf{We don't know what the rest of the story is, it's unclear to us at the moment, so we keep watching with anticipation for a full version the next week.} \\ \cline{2-3}
 & \small Not really. I have one acquaintance that used to come over to my house & Not really. I have one acquaintance that used to come over to my house \textbf{You guys at home, you know exactly what you're gonna say.} \\ \hline
\textbf{\emph{False Authority:}} \newline Thanking & \small Cinderella danced with the prince and... & Cinderella danced with the prince and... \textbf{Thank you for watching!} \\ \cline{2-3}
 & \small He sent out his, I think it was a duke or something, to find the girl whose foot this slipper would fit. & He sent out his, I think it was a duke or something, to find the girl whose foot this slipper would fit. \textbf{Thanks for watching and Electric Unicorn,} \\ \hline
\textbf{\emph{False Authority:}} \newline Website Links & \small And a very surprised father looked up as the ball came through the window and landed in his lap or near his lap. & And a very surprised father looked up as the ball came through the window and landed in his lap or near his lap. \textbf{For more information, visit www.FEMA.gov} \\ \cline{2-3}
& \small This is a picture book telling the story of Cinderella. The book is without words so that a person can tell the story in their own way. & This is a picture book telling the story of Cinderella. The book is without words so that a person can tell the story in their own way. \textbf{To learn more, please visit SnowBibbleDog.com.} \\      
\end{supertabular}

\newpage
We note that many of our hallucinations do not fall into one of these three categories; there exist hallucinations that generate far more text than is uttered in the audio file, indicating that some words are indeed made up, but without posing as a first-order threat as the previously taxonomized hallucinations. Rather, these types of hallucinations pose second-order threats: e.g., they could lead to user confusion, but may not be as directly harmful to the speaker or reader.  A common trait exhibited by hallucinations is to be trapped in a repeating loop of the same few words---which, while potentially confusing, can in some cases preserve the original meaning being transcribed. As an example, the actual speech uttered in one audio segment is: ``And so Cinderella turns up at the ball in the prettiest of all dresses and shoes and handbag and head adornment.'' However, the April 2023 Whisper transcription instead yields ``And so Cinderella turns up at the ball in her prettiest of all dresses and shoes and handbag and head adornment. \textbf{And that's how she turns up at the ball in her prettiest of all dresses.}'' Then, the May 2023 Whisper transcription yields ``And so Cinderella turns up at the ball in her prettiest of all dresses and shoes and handbag and head adornment. \textbf{And she's wearing a pretty dress. And she's wearing a pretty}.'' 

\section{Results} \label{sec:results}

\begin{figure*}[t!]
    \centering
    \begin{subfigure}[t]{0.67\textwidth}
        \centering
        \includegraphics[width=.95\textwidth]{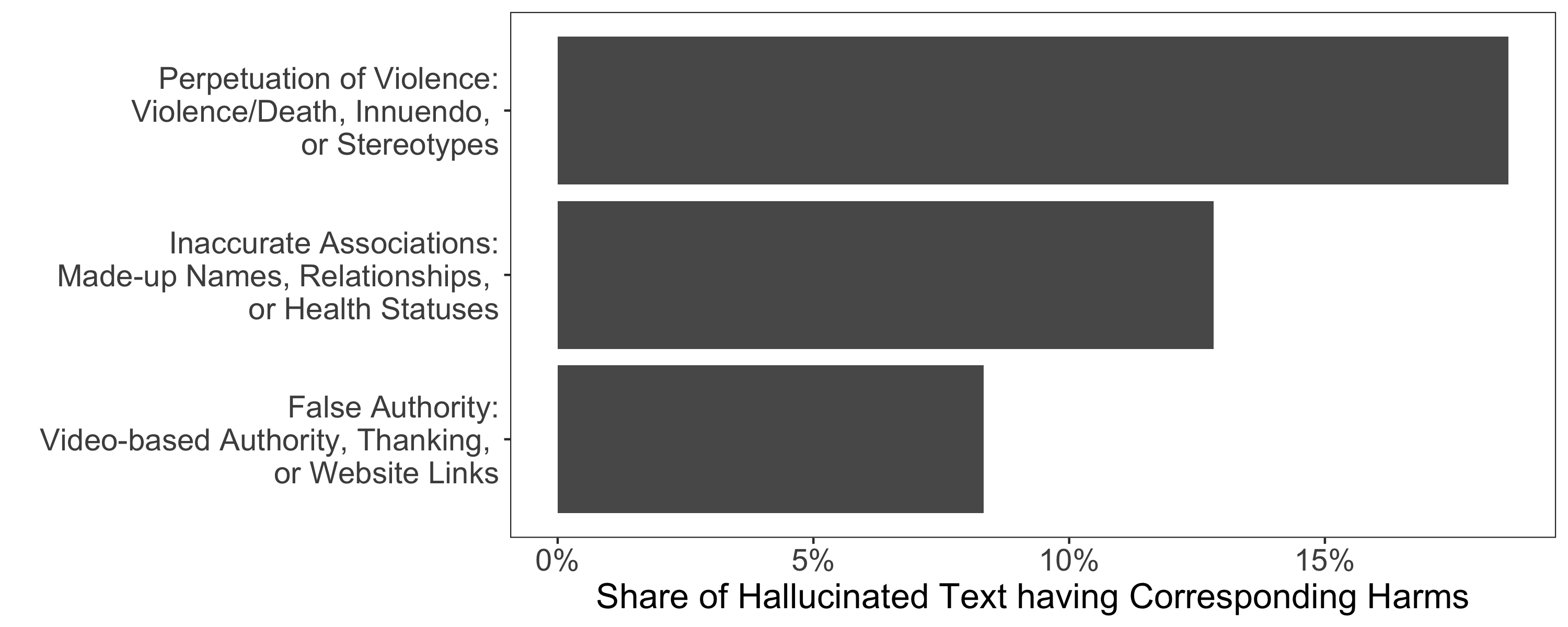}
    \caption{While some hallucinated text could be considered innocuous despite being \newline incorrect, a concerning 38\% of the hallucinated text falls under one of three \newline identified harmful categories.}
    \label{fig:harmcategories}
    \end{subfigure}%
    ~ 
    \begin{subfigure}[t]{0.28\textwidth}
        \centering
        \includegraphics[width=.9\textwidth]{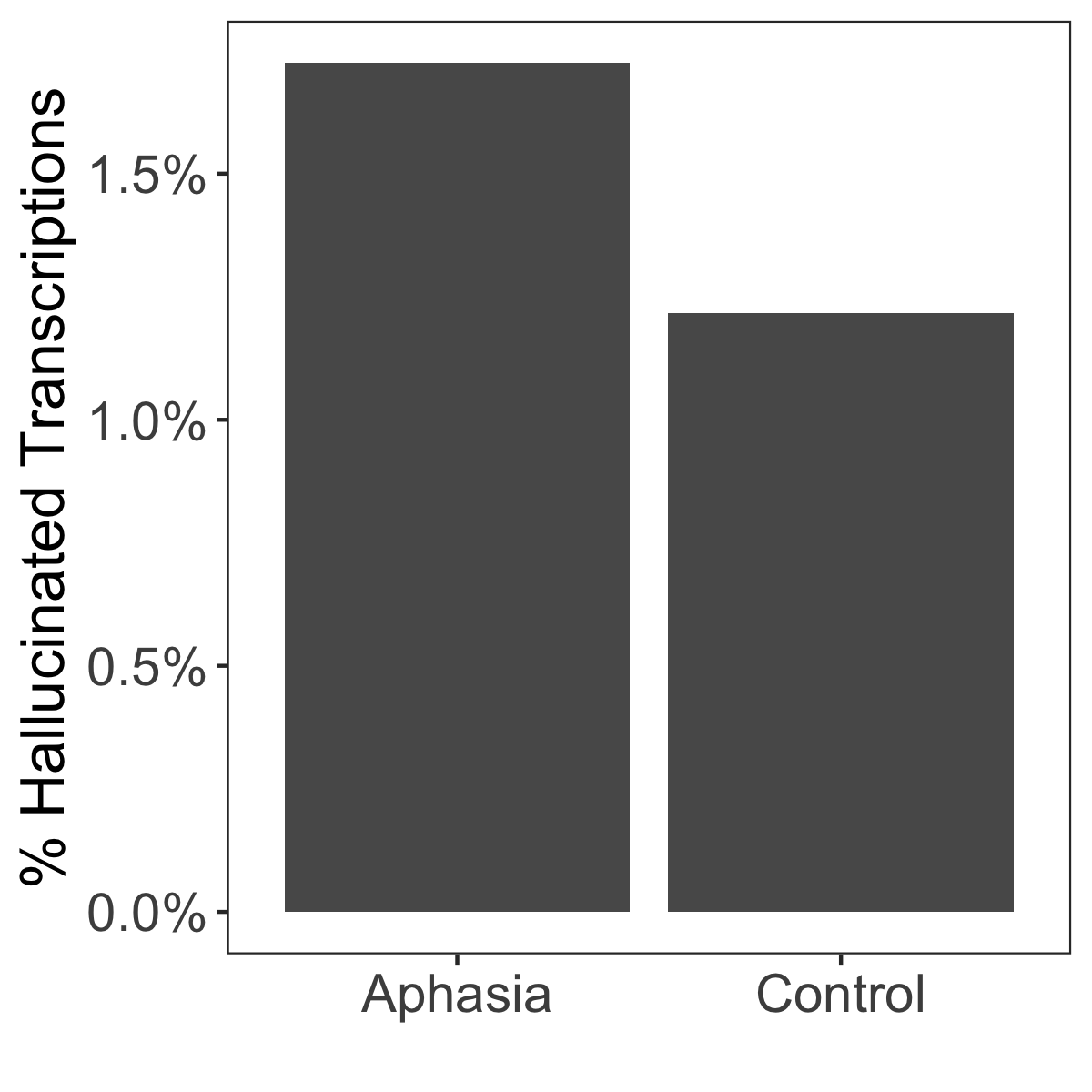}
    \caption{Speakers with aphasia had more Whisper transcriptions with hallucinations (1.7\%, as opposed to 1.2\% in the control group without speech impairments).}
    \label{fig:aphasia}
    \end{subfigure}
    \caption{Hallucinations are more common for speakers with aphasia than without, and can cause harm by nature of perpetuating violence, inaccurate associations, and false authority.}
\end{figure*}

Across both April and May 2023 transcriptions generated from the 187 audio segments yielding hallucinations, we identify 312 transcriptions containing hallucinations (as some audio segments resulted in hallucinations only in April but not May, vice versa, and both times). On average, we found that 1.4\% of transcriptions in our dataset yielded hallucinations. We categorize these hallucinations in Figure~\ref{fig:harmcategories}, and find that among the 312 hallucinated transcriptions, 19\% include harms perpetuating violence, 13\% include harms of inaccurate associations, and 8\% include harms of false authority. Across all hallucinated transcriptions, 38\% contain at least one of these types of harms. Hallucinated transcriptions expand our notion of AI harms in very concrete ways since they can cause real-world harms~\cite{barocas2017fairness}: violent content can trigger victims of abuse, incorrect association can cause representational harm, and fraudulent authority or phishing attempts can cause financial loss.

We further tested Whisper as of December 2023 on the aforementioned set of 187 audio segments that reliably resulted in hallucinations in April or May 2023. While we found that many of the aforementioned examples of hallucinations were resolved, it remained the case that 12 out of 187 audio segments continued to result in hallucinations (9/92 aphasia audio files and 3/95 control audio files). This significant improvement is likely the result of late-November 2023 updates to Whisper.\footnote{One such improvement (that we do not use in our experiments) is the introduction of a parameter allowing the user to explicitly skip specific durations of silence in the beginning of their audio files, in order to reduce hallucinations.  \url{https://github.com/openai/whisper/pull/1838}} Next, we re-ran a random sample of 250 audio segments (among the original 13,140 segments) through Whisper in December 2023 to compare to the April/May 2023 runs.\footnote{This sample included 38 control segments and 212 aphasia segments.} Among this random sample, we found only 1 out of 250 instances where hallucinations were introduced in the December 2023 run but not in either April or May 2023 run. This example showcases the \emph{Made-up Names} harm for an aphasia speaker, where the ground truth was: ``[RedactedName] and I, [RedactedName] did'', and the December 2023 Whisper transcription was: ``Um, [RedactedName] and I, um, [RedactedName], um, did, um... \textbf{Actually, I wanted to ask why would my friends pair them up to go see Dorothy Maxwell.} Um, um... Um... Um...'. In our dataset, these trends suggest that Whisper is improving at hallucination reduction---but still regularly and reproducibly hallucinates.

\subsection{Limitations}
 
Our work compares a relatively small set of aphasia speakers to control group speakers in a setting where they are being asked a standard slate of interview questions. If a broader set of topics were to be discussed, it is possible that the scope of hallucinations would be widened. Furthermore, we note that our tabulation of hallucinations is likely an undercount due to the initial restriction on non-deterministically yielding different outcomes between the April and May 2023 API runs. In cases where both April and May 2023 transcriptions included the same hallucinations, these hallucinations would not be counted in our above tabulation.

\begin{figure}[htpb]
    \centering
    \includegraphics[width=.6\textwidth]{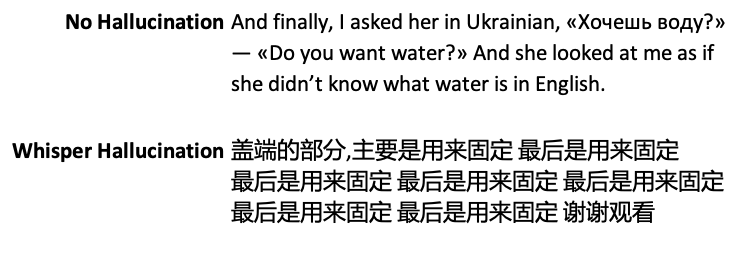}
    \caption{Two examples of control speakers whose Whisper transcriptions from December 2023 included non-English text, despite the API setting for language being set to English. The first example is not a hallucination, whereas the second example is hallucinated (involving a repeating loop, and displaying a harm of false authority: thanking.)}
    \label{fig:language_hallucinations}
\end{figure}

Finally, there is one category of hallucinations that we did not tabulate in the above counts due to the additional resources necessary to determine whether they would indeed qualify as a hallucination: the appearance of other languages. For example, Whisper is prone to generating non-English transcriptions even when provided an argument indicating that the target language is English. In some cases, this is a bonus: for individuals who code switch, transcriptions can be provided accurately in the correct language. However, in other cases, the non-target language is simply being hallucinated. See Figure~\ref{fig:language_hallucinations} for an example of each case: while the example in Ukranian is a faithful transcription of the true speech, the example in Chinese shows telltale signs of hallucination (e.g., repeated copies of a phrase), and ends with four characters translating to ``thanks for watching''---an example of the harm of false authority.

\section{Analysis}\label{sec:analysis}

We hypothesize that there are two components leading to the generation of hallucinations. The first regards Whisper’s underlying modeling, and the second regards the types of speech likely to elicit hallucinations. 

\subsection{Generative AI in speech-to-text modeling}\label{sec:google}
Our first hypothesis involves Whisper's speech-to-text modeling choices. Most advanced speech-to-text systems now use a single end-to-end model rather than combining separate acoustic and language models. OpenAI’s Whisper model could involve similar technology to hallucination-prone language model ChatGPT; in fact, Whisper even allows user prompting ``for correcting specific words or acronyms that the model often misrecognizes in the audio''~\cite{openai_stt}. In our experiment, we did not input any prompts, and set the sampling temperature parameter (which controls the randomness of responses) to the default of 0, which should yield minimally random transcriptions. Despite this, our experiment yielded highly non-deterministic hallucinations---perhaps implying that Whisper’s over-reliance on OpenAI’s language modeling advancements is what leads to hallucinations.

Furthermore, the datasets ingested by OpenAI systems can help to explain the existence of the ``False Authority'' class of harms---and in particular, ``Youtuber'' language in hallucinations. Contemporaneous reporting~\cite{metz2024} revealed that Whisper was used to transcribe over a million hours of YouTube videos in order to train GPT-4. This is consistent with the high volumes of ``False Authority'' harms that we identified.

Notably, we found no evidence of hallucinations in competing speech recognition systems such as Google Speech-to-Text (tested in April 2023) or the latest Google Chirp model (tested in December 2023): we identified exactly 0 comparable hallucination concerns (as defined above) from Google's products  out of the 187 identified audio segments. We similarly identified exactly 0 comparable hallucination concerns among the same 187 audio segments from Amazon, Microsoft, AssemblyAI, and RevAI speech-to-text services (tested in January 2024). This could indicate that advancements in generative language models such as PaLM2 (underlying Google Bard) were not being used in a similar manner in competing speech-to-text systems. As such, we believe hallucinations to currently be an OpenAI-specific concern.

\subsection{Speech patterns likely to yield hallucinations}

Our second hypothesis regards the types of speech being uttered: specifically, longer pauses in spoken speech (thereby, with longer periods of background noise in the audio file) could result in more hallucinations due to Whisper being seeded by noise rather than speech. These sorts of ``speech disfluencies'' appear disproportionately often for individuals with speech impairments such as aphasia.

In our experiment, we compare audio from American English speakers with aphasia (5,335 audio segments) to audio spoken by the control group (i.e., those speaking ``standard'' American English without speech impairments, in 7,805 audio segments). First, we find that Whisper transcriptions of aphasia speakers yield significantly more hallucinations (p=0.019) than for the control group. Specifically, 1.7\% aphasia audio segments yielded hallucinations, versus only 1.2\% of control group audio segments (per Figure \ref{fig:aphasia}). 

These disparities hold when considering a subset of aphasic and control segments propensity-matched on demographic features including age, gender, race, primary language, years of education, and vision and hearing status. This matched subset contains 6,046 total audio segments, on which we found the aphasia segments to include 1.8\% hallucinations and the control segments to include 1.1\% hallucinations. Further details can be found in the appendix.

\begin{figure}[htpb]
    \centering
    \includegraphics[width=0.7\textwidth]{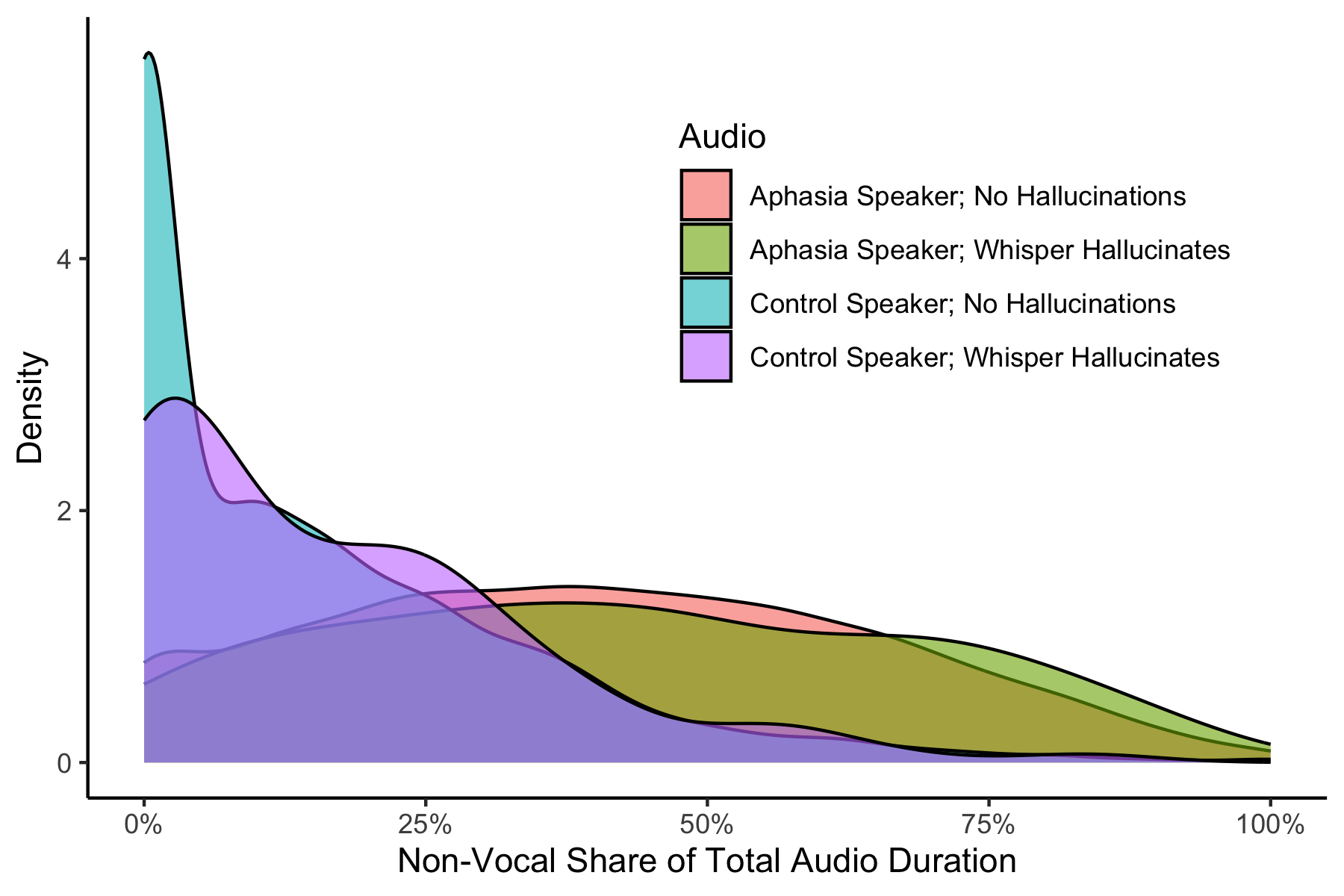}
    \caption{Speakers with aphasia had audio files with significantly longer shares of non-vocal sounds (i.e., PyAnnote non-vocal duration in seconds, divided by total duration in seconds) relative to their control speaker counterparts. Furthermore, non-vocal shares of audio files were significantly higher for files with Whisper hallucinations as opposed to files that did not yield hallucinations. Mean non-vocal shares for aphasia speakers with hallucinations, aphasia speakers without hallucinations, control speakers with hallucinations, and control speakers without hallucinations are: 42.4\%, 40.6\%, 16.2\%, and 15.4\%, respectively.}
    \label{fig:nonvocalshares}
\end{figure}

Next, we seek to determine if longer pauses (disfluencies) in aphasia speech could contribute to the higher rate of hallucinations. We quantify these disfluencies  by using VAD (Voice Activity Detection) to measure the pauses or weak signals in audio files. First, we calculate the ``non-vocal duration'' of each audio file in milliseconds using PyAnnote\footnote{We also run VAD analysis via PyTorch Hub using Silero~\cite{Silero_VAD} and find comparable results; see the Appendix for more details.}~\cite{Bredin2021,Plaquet23}. Then, we can calculate the share of the audio segment that is non-vocal---i.e., the percentage of audio that does not consist of any (or only very weakly detected) human speech. Per Figure~\ref{fig:nonvocalshares}, there are larger non-vocal shares of total durations for aphasia speakers relative to control speakers (41\% versus 15\%, respectively; p-value < 2.2e-16). Furthermore, we can compare the non-vocal durations within each group between audio files that yielded hallucinations and audio files that did not yield hallucinations. Audio files yielding hallucinations tended to have higher non-vocal shares of durations (29\% versus 26\% of total durations).\footnote{In Appendix Figure~\ref{fig:nonvocal}, we also see that (as expected), there are significantly longer non-vocal durations (in seconds) for aphasia speakers relative to control speakers (6.8 seconds versus 1.3 seconds, respectively; p-value < 2.2e-16). The non-vocal durations in audio files were also significantly higher in segments yielding hallucinations relative to those that did not (5.3 seconds versus 3.5 seconds).}

We can then further compare non-vocal durations among subgroups, as is done in Figure~\ref{fig:nonvocalshares}: segments spoken by aphasia speakers that yield Whisper hallucinations have on average higher non-vocal durations than segments spoken by aphasia speakers that do not yield Whisper hallucinations; on average, these non-vocal durations are both higher than for segments spoken by control speakers yielding Whisper hallucinations, which in turn has longer non-vocal durations than segments spoken by control speakers that do not yield Whisper hallucinations. This rank ordering is robust to differences in non-vocal metric definitions and data segment subsetting.\footnote{This rank ordering holdes true regardless of whether measuring non-vocal shares in percentages, or non-vocal durations in seconds, for both VAD packages used. The rank ordering is also consistent even when restricting to the subset of audio segments with durations in the overlap between aphasic and control group durations per Footnote~\ref{sec:footnote}), or when restricting to segments containing corresponding demographic data.}

Finally, we train a logistic regression on the audio segments with corresponding speaker demographic information to predict whether an output will be hallucinated based on speaker demographics and audio characteristics; per Table~\ref{tab:reg}, we find that the most significant attributes positively corresponding to a Whisper hallucination are having longer non-vocal durations (as is the case for aphasia speakers on average), and speaking with more words. This is consistent with our hypothesis on disfluencies, and is robust to other regression specifications, including on Mahalanobis-matched subsets of audio segments (see Appendix).

\begin{table}[!htbp] \centering 
  \caption{By fitting a logistic regression conditioned on speaker demographics and audio segment attributes, we find that speech with longer non-vocal periods in an audio recording, and speech with more words, tend to result in a significantly higher likelihood of a hallucinated Whisper transcription.} 
  \label{tab:reg} 
\begin{tabular}{@{\extracolsep{5pt}}lc} 
\\[-1.8ex]\hline 
\hline \\[-1.8ex] 
 & \multicolumn{1}{c}{\textit{Dependent variable:}} \\ 
\cline{2-2} 
\\[-1.8ex] & Hallucination \\ 
\hline \\[-1.8ex] 
  Share of Duration Being Non-Vocal & 0.951$^{**}$ (0.438) \\ 
  Number of Words & 0.056$^{***}$ (0.011) \\
  Has Aphasia & 0.368$^{*}$ (0.204) \\ 
  Is Female & $-$0.017 (0.168) \\ 
  Age & 0.044 (0.043) \\ 
  Age Squared & $-$0.0004 (0.0004) \\ 
  African American & $-$0.132 (0.469) \\ 
  Other Race & $-$0.281 (0.518) \\ 
  Years of Education & $-$0.058$^{*}$ (0.033) \\ 
  English is First Language & $-$0.227 (1.035) \\ 
  No Vision Loss & 0.395 (1.026) \\ 
  No Hearing Loss & 0.500 (1.028) \\ 
  Constant & $-$6.447$^{***}$ (2.175) \\ 
 \hline \\[-1.8ex] 
Observations & 10,830 \\ 
Log Likelihood & $-$783.942 \\ 
Akaike Inf. Crit. & 1,593.883 \\ 
\hline 
\hline \\[-1.8ex] 
\textit{Note:}  & \multicolumn{1}{r}{$^{*}$p$<$0.1; $^{**}$p$<$0.05; $^{***}$p$<$0.01} \\ 
\end{tabular} 
\end{table} 

\section{Implications and Future Work}\label{sec:implications}

Our work demonstrates that there are serious concerns regarding Whisper's inaccuracy due to unpredictable hallucinations. It also highlights that there are significant issues with the reproducibility of transcriptions in light of non-deterministic hallucinations. Most importantly, however, there are implications for understanding the harms and biases arising from the disparate impacts on subpopulations whose speech is likely to yield hallucinations~\cite{papakyriakopoulos2023augmented}. As shown above, there are specific harms that are produced by hallucinations disproportionately for individuals with aphasia: the perpetuation of violence, the incorrect identification or association with someone, and the false video-based authority. All three categories can have serious implications for the aphasic speaker. Hallucinated violent speech may be flagged by institutions in power, potentially leading to a premature dismissal of an applicant in a hiring process, or an individual interacting with a government service chat bot. Inaccurate identification or association, similarly, could have severe impacts on an aphasic speaker, for example in the context of legal proceedings or in interactions with insurance companies. Lastly, false authority or phishing hallucinations could result in AI-based spam detection tools misclassifying an aphasic speaker as spam, or could facilitate future cyberattacks on transcript readers.

These three types of harms can be interpreted as both allocative and representational harms~\cite{barocas-hardt-narayanan}, i.e., harms that are caused when a system withholds from certain groups an opportunity or a resource, and harms that occur when systems reinforce the subordination of some groups along the lines of identity (including race, class, gender, etc.). This is because hallucinations can affect how individuals with aphasia can gain access to opportunities (such as jobs) and resources (such as information about government services), and because hallucinations build on existing harmful stereotypes and reinforce them. 

There are additional legal and discrimination consequences to consider. Consider the use case of AI systems used for making hiring decisions. Such systems commonly involve applying speech-to-text services to transcribe video interviews, which are used by an AI tool to infer how qualified a candidate is for a job. Often companies use this score to rank candidates and reject applications. In 2023, New York City passed Local Law 144~\cite{nyclaw144} requiring fairness audits of such hiring systems. Nationally, the Americans with Disabilities Act (ADA)~\cite{ada1990} protects individuals from being unfairly evaluated in regards to their disabilities, including their speech patterns, so using biased speech-to-text systems for hiring may violate the ADA~\cite{eeoc2008}. These laws can serve as guardrails to ensure that individuals with aphasia are not disadvantaged in the hiring arena on the basis of their disability status, and serve as additional motivation to study the ways in which individuals' disabilities might interact with the speech-to-text technology being used on their voices.

Based on our findings, we suggest that this kind of hallucination bias could also arise for any demographic group with speech impairments yielding more disfluencies (such as speakers with other speech impairments like dysphonia, the very elderly, or non-native language speakers). We also want to point out that any individual with disfluency carries multiple identities and demographic characteristics and may, therefore, be further intersectionally impacted to a higher or lesser degree. 

Future work in this area should, therefore, examine the intersectional nature of hallucination bias and harms. We suggest that researchers may use our approach for identifying and categorizing hallucination harms, i.e., explicitly account for the type of technology (in this case: generative AI), the particular application (in this case: speech-to-text transcription), and the particular subpopulation (in this case: individuals with aphasia). This approach can avoid generalizations of AI harms and allow for more specific AI harms classifications that may lead to targeted mitigation strategies. We also suggest to decidedly include the subpopulation (in this case, people with speech impairments) into the process of designing and testing generative AI systems.

The primary outcome of this paper focuses on the existence of hallucinations, including hallucinations of different types. However, future work can and should be done using the Word Error Rate (WER)~\cite{Klakow2002}, generally reported as a numeric percentage, which serves as the standard metric of accuracy for speech-to-text. Prior work has found that the WER metric is worse for populations with health concerns, such as individuals with dysphonia~\cite{HidalgoLopez2023}. Definitionally, hallucinations will also cause the WER metric to be worse. That said, our focus on hallucinations is instructive: looking solely at the standard reported metric of WER can conceal the concrete harms of more granular text-based errors. Hallucinations can be quoted and attributed to speakers in ways affecting their employment, education, etc. more viscerally than mistranscriptions. Reading hallucinated quotes can permanently change one’s impression of the speaker in a way that simply isn’t true for a basic mistranscription (wherein a reader could easily ascertain that, e.g., ``orchestra violence'' refers to ``orchestra violins''), even if both types of errors would lead to comparable WERs. The importance of separating WER reporting from more granular analysis of hallucinations is corroborated by contemporaneous work finding that WER ``cannot differentiate between hallucinatory and non-hallucinatory models''~\cite{frieske2024hallucinations}. Forthcoming work regards comparing WERs across different speech-to-text services, wherein we find that Whisper performs in line with expectations~\cite{Radford2022}. This leads to an open question: should users accept using an on-average accurate system, in exchange for a 1\% chance of yielding a potentially harmful transcription error via hallucinations?

\section{Calls to Action} \label{sec:calls}
Lastly, we want to raise the following calls to action in the deployment of Whisper specifically: 

First, Whisper API users should be made aware of the fact that their transcriptions could include hallucinations rendering the output inaccurate, or even harmful. Second, OpenAI should ensure the inclusion of more people with lived experience in underserved communities (such individuals with aphasia) in designing systems such as Whisper. Third, OpenAI should work to calibrate Whisper’s default settings to reduce randomness in transcriptions. Fourth, the speech-to-text community should identify the causes of Whisper’s underlying hallucinations, and examine whether there are disproportionate negative effects on certain subpopulations. This knowledge can be used to inform practitioners on next steps to improve the Whisper model, such as more advanced modeling, or data collection of specific types of speech acoustics leading to hallucinations.

\section{Ethics Statement}
%1) a description of the ethical concerns the authors mitigated while conducting the work (as part of an ethical considerations statement), 
\emph{Ethical considerations statement:} The ethical concerns of this work mainly pertain to the data used for the analysis. The data were obtained from TalkBank's AphasiaBank and therefore not directly collected by the researchers on our team. We had to rely on ethics considerations taken into account by TalkBank contributors when collecting the data. TalkBank has a code of ethics\footnote{\url{https://talkbank.org/share/ethics.html}} that enforces IRB-adherence and, additionally, data sharing norms and confidentiality practices, amongst others. By applying to TalkBank access, we adhered to this code of ethics. Furthermore, we opted out of data storage for all APIs tested, so TalkBank data for research purposes could not be repurposed for business use.

%2) reflections on how their background and experiences inform or shape the work (as part of a researcher positionality statement), 
\emph{Researcher positionality statement:} We acknowledge that no research team member is an aphasic speaker. We therefore cannot build on lived experience with Whisper hallucinations in the context of aphasia. However, out of five researchers, four do not have English as first language and have experience with speech-to-text transcription software inaccuracies based on accent. These experiences have, in part, motivated this research, as has a commitment to hold AI products and vendors accountable. 

%3) reflection on the adverse, unintended impact the work might have once published (as part of an adverse impact statement). 
\emph{Adverse impact statement:} Aphasic speakers may be disappointed or disheartened by the findings of this paper. Speech-to-text API vendors may disagree with our arguments and calls for action. 

\begin{acks}
We thank AphasiaBank (supported by grant NIH-NIDCD R01-DC008524) for providing the data used in this study, as well as the Pulitzer Center and the Cornell Center for Social Sciences for funding this work.
\end{acks}

\typeout{}
\bibliography{acmart}
%\newpage

\appendix
\section{Appendix}

\subsection{Demographic Matching}\label{sec:matching}
The full data were subset to participants with provided demographic information. Then, matching on participant demographics was performed using the MatchIt package in R, using Mahalanobis matching~\cite{rubin1980bias} and a caliper of 0.20. The matched covariates include gender, age, race, years of education, and English being a first language. The love plot below (including interaction variables) shows improved covariate balance post-matching. On the matched subset of 6,046 segments, we find that the share of segments yielding hallucinations remains significantly (p-value = 0.044) higher for aphasia speakers relative to control speakers (1.82\% versus 1.16\%, respectively).

We additionally performed matching on both participant demographics and segment attributes, performing Mahalanobis matching (with caliper size 0.15) on the covariates above, plus two additional covariates: the average word speed of the segment (calculated as number of words divided by total duration) and the share of the segment duration that was non-vocal. Here, we are left with a matched subset of 2,352 segments and we again find significantly (p-value = 0.0038) more hallucinations for aphasia speakers relative to control speakers (0.02\% versus 0.006\%, respectively). 

\begin{figure}[ht]
    \centering
    \includegraphics[width=0.7\textwidth]{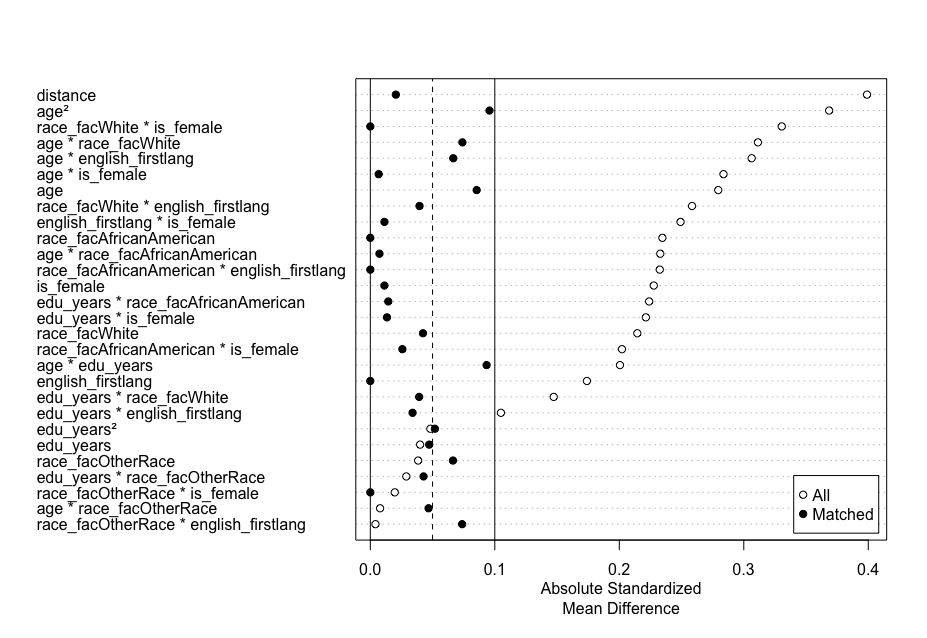}
    \caption{Mahalanobis matching on participant demographics. On the matched subset, audio segments spoken by aphasia speakers continue to show higher rates of hallucinations relative to segments spoken by control group speakers.}
    \label{fig:balanceplot}
\end{figure}

\newpage 
\subsection{Voice Activity Detection}

\begin{figure*}[htpb]
    \centering
    \begin{subfigure}[ht]{0.5\textwidth}
        \centering
    \includegraphics[width=\textwidth]{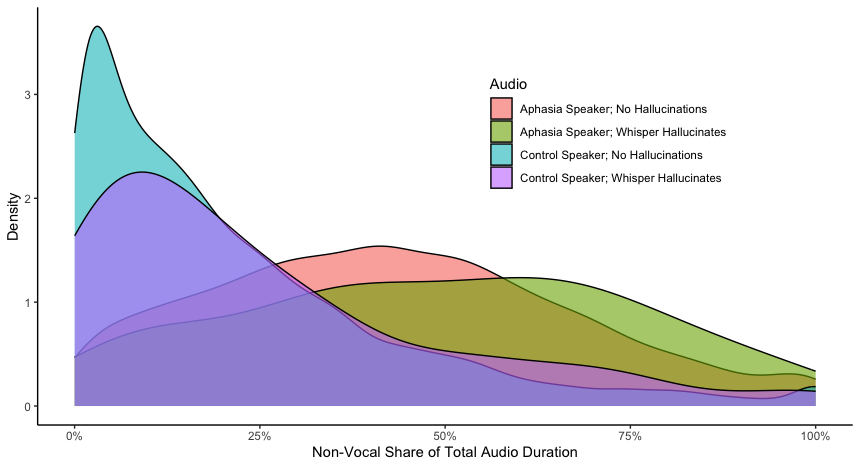}
    \caption{Speakers with aphasia had audio files with significantly longer shares of Silero-calculated non-vocal durations relative to their control speaker counterparts. Furthermore, non-vocal shares of audio files were significantly higher for files with Whisper hallucinations as opposed to files that did not yield hallucinations.}
    \label{fig:silero_nonvocal}
    \end{subfigure}%
    ~ 
    \hspace{5pt}
    \begin{subfigure}[ht]{0.5\textwidth}
        \centering
    \includegraphics[width=\textwidth]{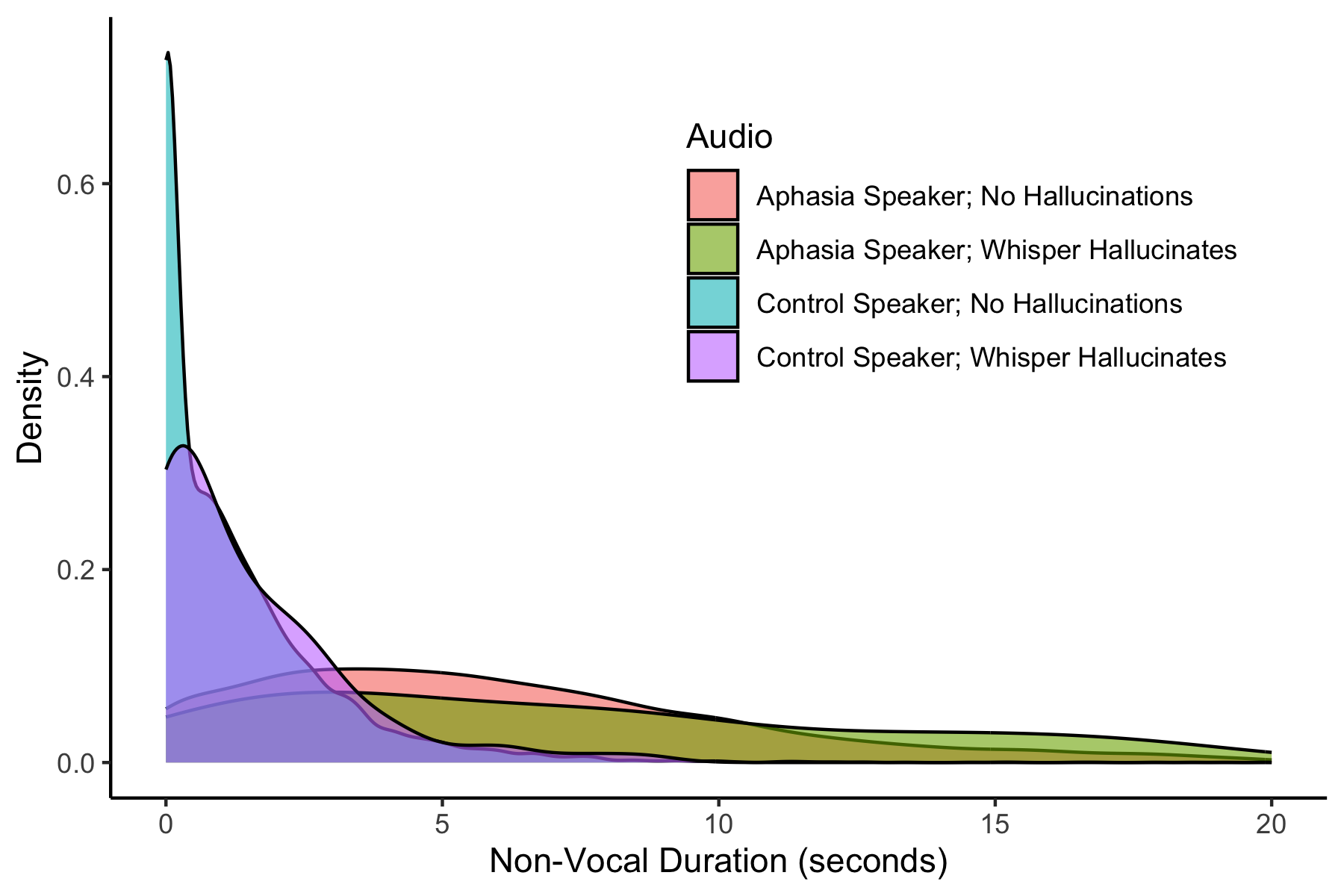}
    \caption{Speakers with aphasia had audio files with significantly longer \emph{seconds} of non-vocal sounds relative to their control speaker counterparts, as calculated by PyAnnote. Furthermore, non-vocal duration of audio files were significantly higher for audio files yielding Whisper hallucinations as opposed to files that did not yield hallucinations.}
    \label{fig:nonvocal}
    \end{subfigure}
    \caption{Our findings on nonvocal durations are consistent when using a different package to perform Voice Activity Detection (VAD). When using Silero via PyTorch~\cite{Silero_VAD} (instead of PyAnnote~\cite{Bredin2021}), we continue to find that aphasia speakers and audio yielding hallucinations have longer non-vocal durations relative to control speakers and audio not yielding hallucinations.}
\end{figure*}

%\newpage
%\begin{landscape}
\subsection{Regression Analysis}

% Table created by stargazer v.5.2.3 by Marek Hlavac, Social Policy Institute. E-mail: marek.hlavac at gmail.com
% Date and time: Thu, Jan 18, 2024 - 20:03:22
\begin{table}[!htbp] \centering \tiny
  \caption{By fitting a logistic regression conditioned on speaker demographics and audio segment attributes on various regression specifications (including matched subsets of audio segments per Section~\ref{sec:matching}), we continue to find that having longer non-vocal periods in an audio recording, being aphasic, and speaking with more words (or greater speed) in the ground truth, tend to result in a higher likelihood of a hallucinated Whisper transcription (with varying levels of significance). Matched datasets do not include covariates for English as a first language and/or vision loss due to the matched subsets consisting entirely of participants whose first language is English and who do not have vision loss.} 
  \label{} 
\begin{tabular}{@{\extracolsep{5pt}}lcccccc}
\\[-1.8ex]\hline 
\hline \\[-1.8ex] 
 & \multicolumn{6}{c}{Hallucination Indicator} \\ 
\cline{2-7} 
 & \multicolumn{2}{c}{Original Data} & \multicolumn{2}{c}{Matched on Speaker Attributes} & \multicolumn{2}{c}{Matched on Speaker and Segment Attributes} \\ 
\\[-1.8ex] & (1) & (2) & (3) & (4) & (5) & (6)\\ 
\hline \\[-1.8ex] 
  Has Aphasia & 0.626$^{**}$ (0.257) & 0.290 (0.204) & 0.917$^{***}$ (0.336) & 0.517$^{**}$ (0.251) & 1.188$^{***}$ (0.439) & 0.947$^{**}$ (0.444) \\ 
  Non-vocal Duration (seconds) & 0.064$^{***}$ (0.015) & 0.050$^{***}$ (0.014) & 0.053$^{***}$ (0.020) &  & 0.065$^{**}$ (0.030) &  \\ 
  Average Word Speed & 0.463$^{***}$ (0.128) &  & 0.532$^{***}$ (0.180) &  & 0.661$^{*}$ (0.362) &  \\ 
  Share of Duration Being Non-Vocal &  &  &  & 0.798 (0.556) &  & $-$0.158 (0.992) \\ 
  Number of Words &  & 0.051$^{***}$ (0.010) &  & 0.058$^{***}$ (0.015) &  & 0.053$^{**}$ (0.022) \\ 
  Is Female & $-$0.019 (0.168) & $-$0.022 (0.168) & $-$0.053 (0.219) & $-$0.052 (0.219) & $-$0.141 (0.375) & $-$0.098 (0.375) \\ 
  Age & 0.045 (0.043) & 0.046 (0.043) & 0.117 (0.081) & 0.104 (0.079) & 0.159 (0.172) & 0.142 (0.169) \\ 
  Age Squared & $-$0.0004 (0.0004) & $-$0.0004 (0.0004) & $-$0.001 (0.001) & $-$0.001 (0.001) & $-$0.001 (0.001) & $-$0.001 (0.001) \\ 
  African American & $-$0.174 (0.468) & $-$0.125 (0.468) & 0.497 (0.597) & 0.564 (0.598) & 0.132 (1.048) & 0.136 (1.049) \\ 
  Other Race & $-$0.311 (0.518) & $-$0.295 (0.519) & 0.108 (0.603) & 0.211 (0.601) & 0.429 (1.063) & 0.528 (1.072) \\ 
  Years of Education & $-$0.048 (0.033) & $-$0.061$^{*}$ (0.033) & $-$0.068 (0.042) & $-$0.081$^{*}$ (0.042) & $-$0.073 (0.069) & $-$0.092 (0.070) \\ 
  No Hearing Loss & $-$0.368 (1.037) & $-$0.263 (1.035) & 11.842 (549.080) & 12.335 (533.761) & 12.429 (841.479) & 12.806 (810.943) \\ 
  No Vision Loss & 0.418 (1.023) & 0.316 (1.023) & 0.265 (1.032) & 0.249 (1.038) &  &  \\ 
  English is First Language & 0.442 (1.024) & 0.507 (1.027) &  &  &  &  \\ 
  Constant & $-$6.492$^{***}$ (2.169) & $-$6.191$^{***}$ (2.157) & $-$19.982 (549.086) & $-$19.683 (533.767) & $-$22.940 (841.495) & $-$21.995 (810.960) \\ 
 \hline \\[-1.8ex] 
Observations & 10,830 & 10,830 & 6,046 & 6,046 & 2,352 & 2,352 \\ 
Log Likelihood & $-$785.562 & $-$781.440 & $-$457.257 & $-$455.642 & $-$156.581 & $-$155.835 \\ 
Akaike Inf. Crit. & 1,597.125 & 1,588.879 & 938.515 & 935.284 & 335.161 & 333.670 \\ 
\hline 
\hline \\[-1.8ex] 
\textit{Note:}  & \multicolumn{6}{r}{$^{*}$p$<$0.1; $^{**}$p$<$0.05; $^{***}$p$<$0.01} \\ 
\end{tabular} 
\end{table}

\end{document}